# Encoding Longer-term Contextual Multi-modal Information in a Predictive Coding Model

Junpei Zhong*[†], Tetsuya Ogata*[‡], Angelo Cangelosi[†]
*National Institute of Advanced Industrial Science and Technology (AIST), Aomi 2-3-26, Tokyo, Japan
Email: joni.zhong@aist.go.jp
[†]Centre for Robotics and Neural Systems, Plymouth University, Plymouth, UK
[‡] Lab for Intelligent Dynamics and Representation, Waseda University, Tokyo, Japan

*Abstract*—Studies suggest that within the hierarchical architecture, the topological higher level possibly represents a conscious category of the current sensory events with a slower changing activities. They attempt to predict the activities on the lower level by relaying the predicted information. On the other hand, the incoming sensory information corrects such prediction of the events on the higher level by the novel or surprising signal. We propose a predictive hierarchical artificial neural network model that examines this hypothesis on neurorobotic platforms, based on the AFA-PredNet model. In this neural network model, there are different temporal scales of predictions exist on different levels of the hierarchical predictive coding, which are defined in the temporal parameters in the neurons. Also, both the fast- and the slow-changing neural activities are modulated by the active motor activities. A neurorobotic experiment based on the architecture was also conducted based on the data collected from the VRep simulator.

## I. INTRODUCTION

Predictive coding (PC) [1, 2, 3, 4] asserts that our sensorimotor loop works as a predictive machine. In this predictive machine, it attempts to minimize the difference between the posterior estimation and the truth from its perception, by changing its internal learning model ("perceptual inference" (see also [5] and [6]) or by the action execution ("active inference", see also [7] and [8]). Additionally, because of the integrative property of both perception and action, perceiving the world (perceptual inference) and acting on it (active inference) can be regarded as two aspects with the same aim: to minimize the prediction error.

The integrative process of model adjustment follows a bi-directional learning mechanism on each level of our hierarchical brain. It is suggested that within the hierarchical architecture, the topological higher level in the brain areas infer the prediction on the lower areas with a slower changing activities [9, 10]. This is done by its subsets of such prediction representations are transmitted to the lower levels to predict the upcoming faster neural activities on the lower level. For instance, areas on the higher-level of our brain learn multiple world models and act as prior to explain the best descriptions of the upcoming percept. This continual process acts as an "explain away" function (e.g. [11, 12]): the explanation on the higher-level offers the best parameters to predict the most likely causes of the sensory data on the lower levels, which explain away the other models. Such hierarchical function can be realised by the interaction of neural oscillations in different time-scales, which encode different temporal parameters of the world models.

Therefore, the higher level representation in a hierarchical model may physically represent the contextual information based on the understanding of the upcoming world model for prediction, As such, the internal world model on the higher level has to be shaped by the statistical structure of the error. Based on this hypothesis, we suggest that the concept of time-scales should also be implemented in the internal models of the PC framework as well.

## II. RELATED WORKS

The difference of the temporal scales of prediction results in different cognitive functions in embodied internal models. Some of the previous research focused on the short-term predictive function of the internal model. In most cases, such short-term prediction can act a compensation function of the sensorimotor integration (e.g. [13, 14, 15, 16]). Based on the PC framework, the PredNet model [17] is considered to be the first deep learning model that can be utilised in solving a real application. Specifically, the model uses the error between the predicted image and the real image as an input in the bottom-up stream, which strictly follows the concept of PC. In its experiment, the next video frame of the autonomous driving stream could be predicted. To solve this problem, a recent work [18] proposed a model called AFA-PredNet which integrates both motor action and perception in the PC framework. In this network, the motor action is used as attention model for the prediction from a couple of recurrent networks. However, the long-term prediction based on the understanding of the world model is still missing in both the PredNet and the AFA-PredNet models.

Indeed, when we think about the predictive functions in biological brains, there are no explicit boundaries between the short- and mid-term prediction and the long-term predictive: the short-term prediction is based on a long-term understanding and prediction of the world. For instance, [19] and [20] studied how to apply internal model to control the actual motor actions, mostly focusing on the predictive control of a motor action. [21] extended these models to imitation learning of the sensorimotor behaviours. The long-term planning behaviours can also emerge from internal simulation where the prediction occurs constantly (e.g. [22, 23]).

Specifically, while we consider the pre-symbolic representation as a understanding of the context in the long-term prediction, it can be acted as a modality of long-term prediction too, which is learnt in a unsupervised way. From this perspective, [24] reported an embodied experiment in which an association between the semantic meaning and the sensorimotor behaviours emerges by a recurrent architecture called Recurrent Neural Network with Parametric Bias Units (RNNPB). Based on the extension of this network, [25] discovered that the semantic representation about the object movements and object features also emerge in a recurrent neural network. Specifically, the network is able to predict the next probable position of the object movement, while to pre-symbolic representation is given.

If we regard the unification of different time-scales in a single predictive model with artificial recurrent connections, experiments based on the Multiple Timescale Neural Network (MTRNN) [26] offers an explanation from the view of the non-linear dynamical system for such phenomena. It can be regarded as another extended version of the RNNPB. The neurons on the higher-level of the MTRNN are with slower-changing neural activities, which modulates the neural activities on the lower-levels by the similar roles of the bias inputs. Thus, the whole network is able to work as a number of non-linear dynamic functions as a similar role of bifurcation. While the model is used to learn the temporal sequences such as the sensorimotor information of the robots, the model is able to represent different spatio-temporal embodiment scales of sensorimotor information, such as the language learning [27, 25] and object features/movements [28]. Similar concept of multiple time-scales has also be applied in Gated Recurrent Units for automatically context extraction [29, 30].

The multiple time-scales concept can also be extended in different modalities. For instance, the multiple spatio-temporal scales RNN (MSTRNN) [31] integrates the MTRNN and convolutional neural networks [32, 33], where both the spatial and temporal information are connected and asscocated on the higher level, where slower changing neurons represent the sensorimotor behaviours. The slower changing units on the higher level also makes the dynamics of the model easier to be interpreted, examined and changed. But unfortunately, neither MTRNN nor MSTRNN cannot be considered as a PC model, as that they do not have an explicit input from the error (i.e.difference) between the prediction and the original values. On the other hand, compared with MSTRNN, the PredNet [17] follows the definition of PC while using the difference as inputs on each layer. And it also uses the convolutional network to capture the local features of the visual streams. But the PredNet builds the temporal prediction in the top-down perception part, which makes the model more biological plausible.

Building the PC embodied model with the concept of multiple time-scales would be beneficial for both engineering and cognitive studies. Firstly, it follows the results from the brain and cognitive studies that different response times while the neurons react to conscious/unconscious prediction. Second, the slower changing neurons in such a model would be easier for us to control and examine the dynamical behaviours of the model or the embodied systems. This is the main motivation why we are proposing for a novel action modulated PC model with multiple time scales.

## III. THE MODEL

The proposed MTA-PredNet (Multiple Time-scale Action modulated PredNet) is shown in Fig. 1. In general, the MTA-PredNet is functionally organized as an integration with two networks: the left part is equivalent to a generative recurrent network, while the right part is a standard convolutional network.

In terms of architecture, it is similar as AFA-PredNet [18].
1) There are a number of recurrent neural networks. (e.g. Convolutional LSTM) on each level of the model, which learn different possibilities of the prediction (a generative unit, *GU*, green))
2) The input of the motor action is used as an additional signal for the modulation of the prediction (the motor modulation unit, *MM*, grey). Specifically, it acts as an attention mechanism for the prediction from the upper level (top-down prediction);
3) The convolutional network in the bottom-up part capture the feature of the error on each level, (the discriminative unit, *DU*, blue);
4) The difference of the updating rate on different levels of the architecture determine different representation of the sptio-temporal properties of the sensorimotor behaviours (the error unit, *ER*, red).

The generative unit, *GU*, is usually a recurrent network that generates a prediction of the next time-step from the current input. Here, the convolutional LSTM [34, 35] is employed to generate the local feature prediction in the image. We employ a number of independent recurrent units on one layer of the *GU* unit so that the different possibilities of the prediction given the motor action input can be memorized and predicted. Such memories are also determined by the time-scales we mention later, which produces the prediction given the contexts of different time-scales. During training with various action-perception pairing, each of these units implicitly memorizes different possibilities of the prediction (e.g. the moving direction) with respect to the motor action in a unsupervised way.

The neural functions on each neural unit can be found in Eq. III. Although the main architecture of the MTA-PredNet is the same as AFA-PredNet, the most important feature is that in the neural function of the generative unit (Eq. 4), the generated output is determined not only by the current neural status, but also its previous status. The fraction of the output is determined by the temporal parameter $\tau$.

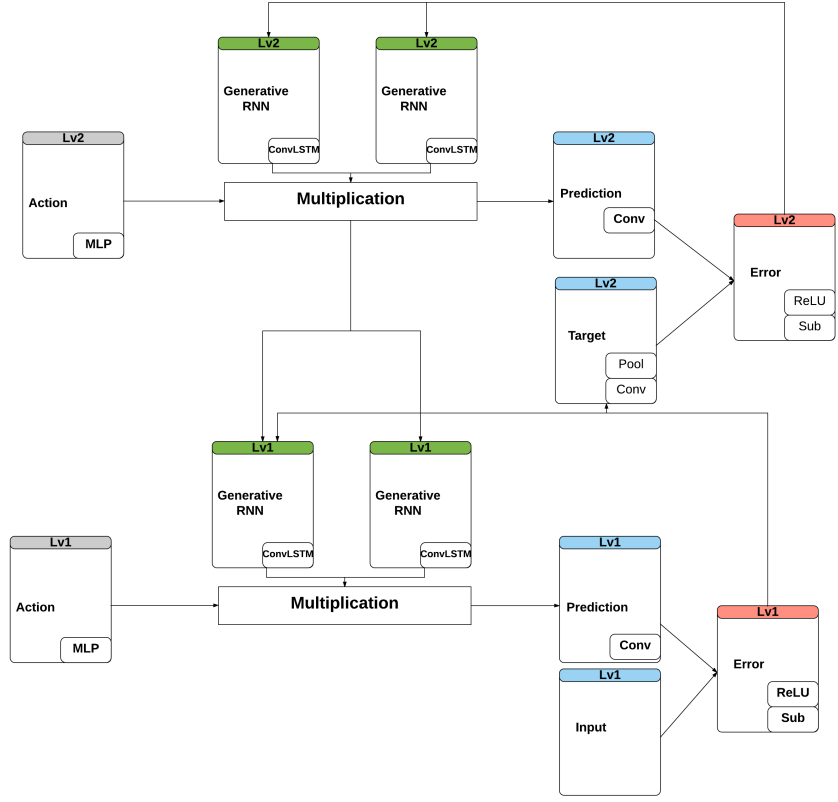

Fig. 1: A 2-layer AFA-PredNet

$$X_l(t) = \begin{cases} i(t), & \text{if } l = 0, \\ MAXPOOL(f(Conv(E_{l-1}(t)))), & \text{if } l > 0 \end{cases} \qquad (1)$$

$$\hat{X}_l(t) = f(Conv(R_l(t))) \qquad (2)$$

$$E_l(t) = [f(X_l(t) - \hat{X}_l(t)); f(\hat{X}_l(t) - X_l(t))] \qquad (3)$$

$$R_l^d(t) = (1 - \frac{1}{\tau})R_l^d(t) + \frac{1}{\tau}ConvLSTM(E_l(t-1), R_l(t-1), DevConv(R_{l+1}(t))) \qquad (4)$$

$$R_l(t) = MLP(a(t)) \times R_l^d(t) \qquad (5)$$

where $f(\cdot)$ is an activation function of the neurons, which we apply the ReLu function to ensure a faster learning in back-propagation, $X(\cdot)_l^t$ is the neural representation of the level $l$ at time $t$. The representation on the *EL* layer $l$ is $E(\cdot)_l$. The $MAXPOOL$, $Conv$, $ConvLSTM$ and $MLP$ are the corresponding neural algorithms. Specifically, to realize the time scale concept, Eq. 4 indicates that the predicted information in the *GU* unit should consider the previous state of the ConvLSTM outputs as well as the current output. This is determined by the time parameter $\tau$.

The overall algorithm for learning a whole sequence is showed in Algorithm 1.

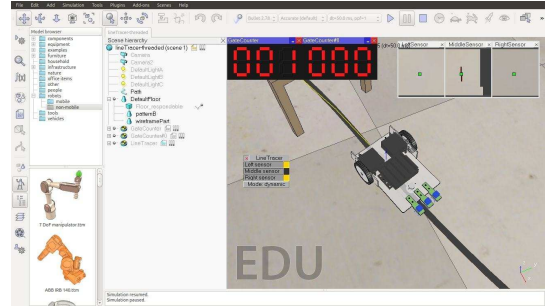

Fig. 2: Data Collected from VRep Simulation

## IV. CASE ANALYSIS

In this section, the performance of the network as well as the analysis of the neural activities will be conducted in

**Algorithm 1:** MTA-PredNet Computation

```
Data: i(t)&a(t) ∈ data
while error > threshold or
 iteration > maximum_iteration do
    for t ← 0 to T do
        for l ← 0 to L do
            if l == L then
                R_l^d(t) = (1 − 1/τ)R_l^d(t − 1) + 1/τ ·
                ConvLSTM(E_l(t − 1), R_l(t − 1));
            else
                R_l^d(t) = (1 − 1/τ)R_l^d(t − 1) + 1/τ ·
                ConvLSTM(E_l(t − 1), R_l(t −
                1), DevConv(R_{l+1}(t)));
            end
            R_l(t) = MLP(a(t)) × R_l^d(t);
        end
        /* Generative (top-down) Process
           */
        for l ← L to 0 do
            X̂_l(t) = f(Conv(R_l(t))); E_l(t) =
            [f(X_l(t) − X̂_l(t)); f(X̂_l(t) − X_l(t));
            /* Discriminative
               (bottom-up) Process     */
        end
    end
end
```

a mobile robot experiment. We recorded a data-set from a robot simulation about the line tracer robot car from the VRep simulator [36]. In this simulation (Fig. 2), the robot equips three vision sensors as well as three Line Finder sensors. With these sensors, the robot was able to adjust the velocities of its wheels to follow the line on the ground. Using VRep, we were also able to record the wheel velocity data and the camera data to train the network. To gather the data, we captured the grey-scale images with size of $8 \times 12$ pixels from the middle vision sensor every 0.02s.

A three-layer MTA-PredNet was used for training the sequence of both motor action vectors (i.e. the velocities of the wheels) and images, with the Adam optimizer [37]. Three different values of $\tau$ were applied in three different layers. With a larger $tau$ on the upper levels, it indicates slower neural activities would be expected. Compared with the $\tau$ values selected in MTRNN works (e.g. [26, 28]), a much smaller $\tau$ values are chosen, because the LSTM networks performs longer term memories by themselves. The parameters are shown in the table:

| Parameters | Value |
|---|---|
| $\tau_0$ | 1.0 |
| $\tau_1$ | 1.3 |
| $\tau_2$ | 2.0 |
| Kernel | $3 \times 3$ |
| Padding | 1 |
| Pooling | $2 \times 2$ |

TABLE I: parameters

Fig. 3 and Fig. 4 show the comparison between the samples of the original and the predicted images.

We further visualise the neural activities on different layers to examine how time parameters $\tau$ affects the representation. Corresponding to the prediction samples, the internal representations of the prediction on the 1st $GU$ of each layer are shown (Figs. 5, 6 and 7), from which we can observe the predicted image on the higher-level (Fig. 7) remains steady during almost the whole movement of the robot compared with other two layers. A demo of the experiment can be found in[1].

## V. CONCLUSION

The top-down prediction in the PC framework may occurs based on the longer term understanding representing the contextual multi-modal information. As a few neuroscience studies have suggested the temporal difference in neural activities can be found in the hierarchical brain areas, the multiple time-scales concepts have been applied in an embodied PC model, the PredNet model. Specifically, the higher-level encodes the slowly changing information of both perception and action, indicating the understanding of the full sensorimotor event.

At the next stage, we will examine the network performance in details and with more robot experiments. Also, it would be interesting to explore the interaction between the short- and long-term prediction in the sense of the neural representation. And how such interaction emerges from the embodied interaction.

## ACKNOWLEDGEMENT

The research was partially supported by New Energy and Industrial Technology Development Organization (NEDO). A Pytorch implementation of MTA-PredNet can be found on Github[2]

---

[1] https://youtu.be/4w7RqeU42XY
[2] https://github.com/jonizhong/mta_prednet.git

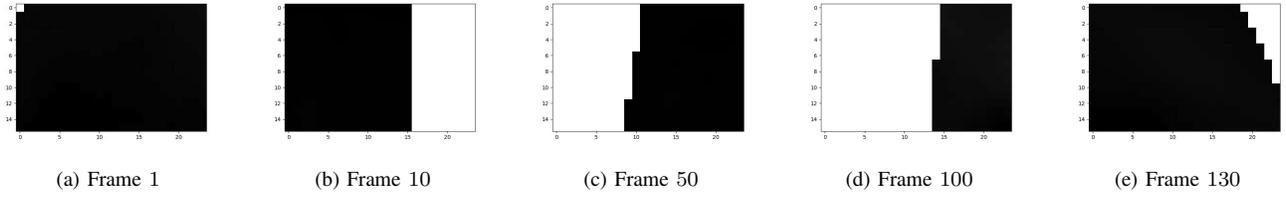

(a) Frame 1  (b) Frame 10  (c) Frame 50  (d) Frame 100  (e) Frame 130

Fig. 3: Image Samples from the Middle Vision Sensor

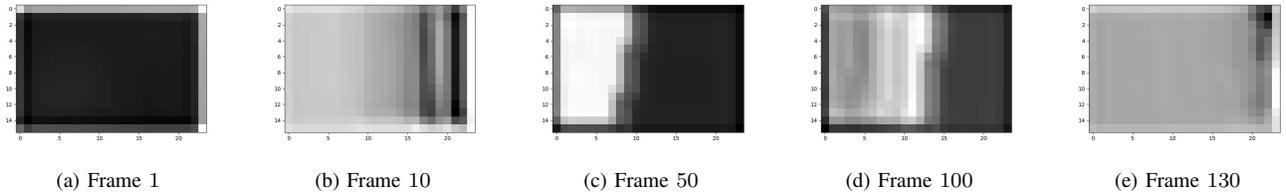

(a) Frame 1  (b) Frame 10  (c) Frame 50  (d) Frame 100  (e) Frame 130

Fig. 4: Predicted Images after Training

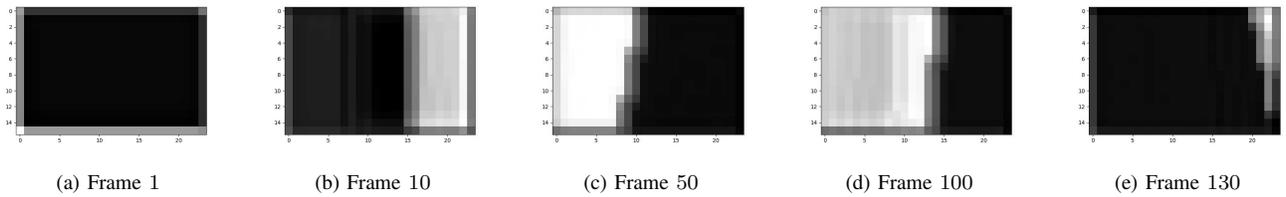

(a) Frame 1  (b) Frame 10  (c) Frame 50  (d) Frame 100  (e) Frame 130

Fig. 5: Image generated from the 1st $GU$ output (Layer 0), $\tau = 1.0$

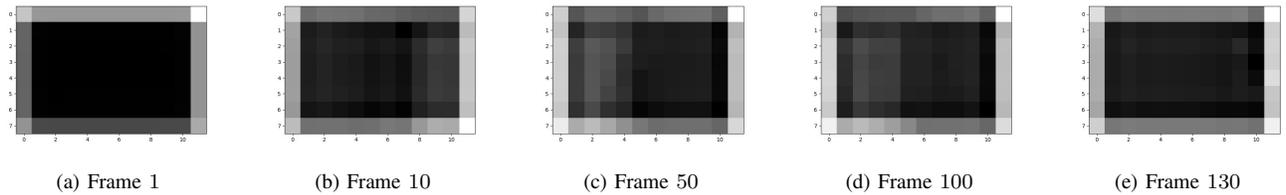

(a) Frame 1  (b) Frame 10  (c) Frame 50  (d) Frame 100  (e) Frame 130

Fig. 6: Image generated from the 1st $GU$ output (Layer 1), $\tau = 1.3$

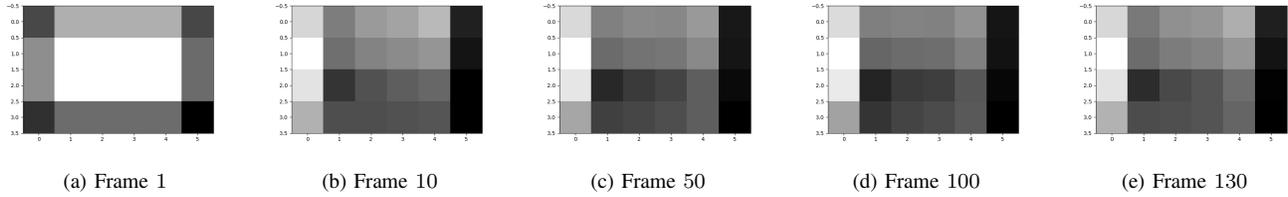

(a) Frame 1    (b) Frame 10    (c) Frame 50    (d) Frame 100    (e) Frame 130

Fig. 7: Image generated from the 1st $GU$ output (Layer 1), $\tau = 2.0$